\documentclass[]{article}
\usepackage[letterpaper]{geometry}
\usepackage{mtsummit2021}
\usepackage{times}
\usepackage{natbib}
\usepackage{url}
\usepackage{latexsym}
\usepackage{layout}
\usepackage[dvipsnames]{xcolor}

\usepackage{booktabs}
\usepackage{graphicx}
\usepackage{lipsum}
\usepackage{amsmath}
\usepackage{breakurl}
\usepackage{hyperref}

\usepackage{multicol}
\usepackage{xspace}

\newcommand{\en}{$en$}
\newcommand{\af}{$af$}
\newcommand{\sw}{$sw$}
\newcommand{\yo}{$yo$}
\newcommand{\kn}{$kn$}
\newcommand{\my}{$my$}
\newcommand{\fr}{$fr$}

\newcommand{\nep}{$ne$}

\newcommand{\ssnmt}{SSNMT}

\newcommand*{\yoruba}{Yor\`ub\'a\xspace}

\begin{document}

\title{\bf Integrating Unsupervised Data Generation \\
into Self-Supervised Neural Machine Translation \\
for Low-Resource Languages}

\author{\name{\bf Dana Ruiter} \hfill  \addr{druiter@lsv.uni-saarland.de}\\
        \name{\bf Dietrich Klakow} \hfill \addr{dietrich.klakow@lsv.uni-saarland.de}\\
        \addr{Spoken Language Systems Group, Saarland University, Germany}
\AND
      \name{\bf Josef van Genabith} \hfill \addr{josef.van\_genabith@dfki.de}\\
        \addr{DFKI GmbH \& Saarland University, Saarland Informatics Campus, Saarbr\"ucken, Germany}
\AND
      \name{\bf Cristina Espa\~{n}a-Bonet} \hfill \addr{cristinae@dfki.de} \\
        \addr{DFKI GmbH, Saarland Informatics Campus, Saarbr\"ucken, Germany}

}

\maketitle
\pagestyle{empty}

\begin{abstract}
\vspace{4mm}
For most language combinations, parallel data is either scarce or simply unavailable. To address this, unsupervised machine translation (UMT) exploits large amounts of monolingual data by using synthetic data generation techniques such as back-translation and noising, while self-supervised NMT (\ssnmt) identifies parallel sentences in smaller comparable data and trains on them. To date, the inclusion of UMT data generation techniques in \ssnmt\ has not been investigated.
We show that including UMT techniques into \ssnmt\ significantly outperforms \ssnmt\ and UMT on all tested language pairs, with improvements of up to $+4.3$ BLEU, $+50.8$ BLEU, $+51.5$ over \ssnmt, statistical UMT and hybrid UMT, respectively, on Afrikaans to English. We further show that the combination of multilingual denoising autoencoding, \ssnmt\ with backtranslation and bilingual finetuning enables us to learn machine translation even for distant language pairs for which only small amounts of monolingual data are available, e.g. yielding BLEU scores of $11.6$ (English to Swahili).

\end{abstract}

\section{Introduction}
\label{sec:Intro}

Neural machine translation (NMT)
achieves high quality translations when large amounts of parallel data are available \citep{WMT:2020}. Unfortunately, for most language combinations, parallel data is non-existent, scarce or low-quality. To overcome this, unsupervised MT (UMT) \citep{lampleEtAl:EMNLP:2018, ren2019unsupervised, artetxe2019effective} focuses on exploiting large amounts of monolingual data, which are used to generate synthetic bitext training data via various techniques such as back-translation or denoising.
Self-supervised NMT (\ssnmt) \citep{ruiter-etal-2019-self} learns from smaller amounts of \textit{comparable} data --i.e. topic-aligned data such as Wikipedia articles-- by learning to discover and exploit similar sentence pairs. However, both UMT and \ssnmt\ approaches often do not scale to low-resource languages, for which neither monolingual nor comparable data are available in sufficient quantity \citep{guzman-etal-2019-flores,espanaEtAl:WMT:2019,marchisio2020does}.
To date, UMT data augmentation techniques have not been explored in \ssnmt. However, both approaches can benefit from each other, as $i)$ \ssnmt\ has strong internal quality checks on the data it admits for training, which can be of use to filter low-quality synthetic data, and $ii)$ UMT data augmentation makes monolingual data available for \ssnmt. 

In this paper we explore and test the effect of combining UMT data augmentation with \ssnmt\ on different data sizes, ranging from very low-resource ($\sim66k$ non-parallel sentences) to high-resource ($\sim20M$ sentences). We do this using a common high-resource language pair (\en--\fr), which we downsample while keeping all other parameters identical. We then proceed to evaluate the augmentation techniques on different truly low-resource similar and distant language pairs, i.e. English (\en)--\{Afrikaans (\af), Kannada (\kn), Burmese (\my), Nepali (\nep), Swahili (\sw), \yoruba\ (\yo)\}, chosen based on their differences in typology (\emph{analytic}, \emph{fusional}, \emph{agglutinative}), word order (\emph{SVO}, \emph{SOV}) and writing system (\emph{Latin}, \emph{Brahmic}). We also explore the effect of different initialization techniques 
for \ssnmt\ in combination with finetuning.

\section{Related Work}
\label{s:sota}

Substantial effort has been devoted to muster training data for \textbf{low-resource NMT}, e.g. by identifying parallel sentences in monolingual or noisy corpora in a pre-processing step \citep{artetxe2018margin,chaudhary-EtAl:2019:WMT, schwenk2019wikimatrix} and also by leveraging monolingual data into supervised NMT e.g. by including autoencoding \citep{currey-etal-2017-copied} or language modeling tasks \citep{gulcehre2015using, ramachandran-etal-2017-unsupervised}.  Low-resource NMT models can benefit from high-resource languages through transfer learning \citep{zoph-etal-2016-transfer}, e.g. in a zero-shot setting \citep{johnson2016google}, by using pre-trained language models \citep{lample2019cross,kuwanto2021low}, or finding an optimal path for pivoting through related languages \citep{leng2019unsupervised}.

\textbf{Back-translation} often works well in high-resource settings \citep{bojar-tamchyna-2011-improving, sennrich-etal-2016-improving, Karakanta2018}.
NMT training and back-translation have been used in an incremental fashion in both unidirectional \citep{hoang-etal-2018-iterative} and bidirectional systems \citep{zhang2018joint, niu2018bidirectional}. 

\textbf{Unsupervised NMT} \citep{lampleEtAl:ICLR:2018,artetxeEtAl:ICLR:2018,yangEtAl:2018} applies bi-directional back-translation in combination with denoising and multilingual shared encoders to learn MT on very large monolingual data. This can be done multilingually across several languages by using language-specific decoders \citep{sen-etal-2019-multilingual}, or by using additional parallel data for a related pivot language pair \citep{li-etal-2020-reference}.
Further combining unsupervised neural MT with phrase tables from statistical MT leads to top results  \citep{lampleEtAl:EMNLP:2018,ren2019unsupervised, artetxe2019effective}.
However, unsupervised systems fail to learn when trained on small amounts of monolingual data \citep{guzman-etal-2019-flores}, when there is a domain mismatch between the two datasets \citep{kim-etal-2020-unsupervised} or when the languages in a pair are distant \citep{koneru-etal-2021-unsupervised}. Unfortunately, all of this is the case for most truly low-resource language pairs.

\textbf{Self-supervised NMT} \citep{ruiter-etal-2019-self} jointly learns to extract data and translate from comparable data and works best on 100s of thousands of documents per language, well beyond what is available in true low-resource settings.

\section{UMT-Enhanced SSNMT}
\label{s:architecture}

\begin{figure}
    \centering
    \includegraphics[width=0.4\columnwidth]{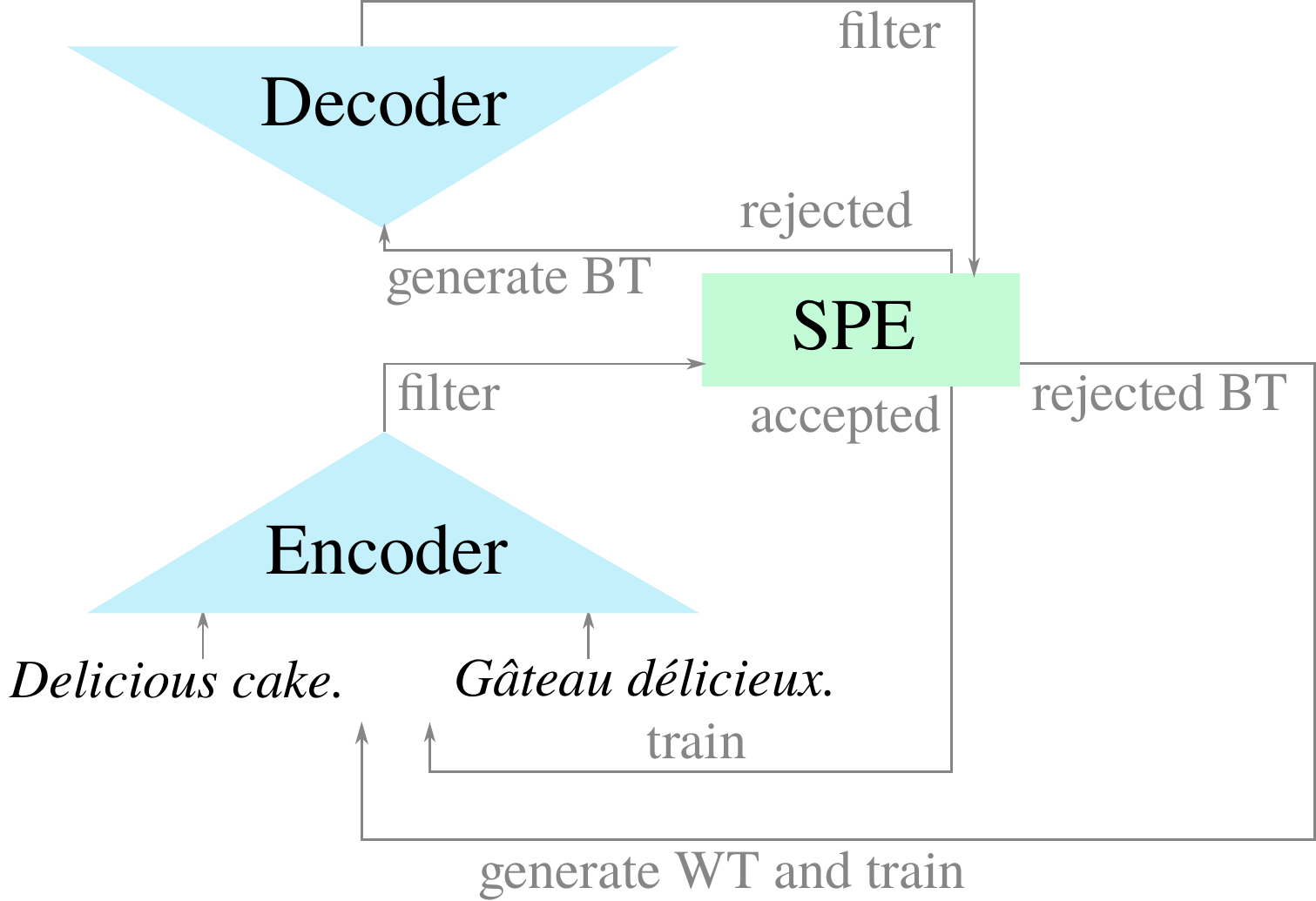}
    \caption{UMT-Enhanced \ssnmt\ architecture (Section \ref{s:architecture}).
    }
    \label{fig:architecture}
\end{figure}

\ssnmt\  jointly learns MT and extracting similar sentences for training from comparable corpora in a loop on-line. Sentence pairs from documents in languages $L1$ and $L2$ are fed as input to a bidirectional NMT system $\{L1, L2\}\rightarrow\{L1, L2\}$,
which filters out non-similar sentences after scoring them with a similarity measure calculated from the internal embeddings. 

\paragraph{Sentence Pair Extraction (SPE):}  Input sentences $s_{L1}$ $\in$ $L1$,  $s_{L2}$ $\in$ $L2$, are represented 
by the sum of their word embeddings and 
by the sum of the encoder outputs, and scored using the margin-based measure introduced by \citet{artetxe2018margin}. If a pair ($s_{L1}, s_{L2}$) is top scoring for both language directions \emph{and} for both sentence representations, it is accepted for training, otherwise it is filtered out. This is a strong quality check and equivalent to \emph{system P\,} in \citet{ruiter-etal-2019-self}. A \ssnmt\ model with SPE is our \textbf{baseline (B)} model.

\paragraph{} Since most possible sentence pairs from comparable corpora
are non-similar, they are simply discarded. In a low-resource setting,
this potentially constitutes a major loss of usable monolingual information. To exploit sentences that have been rejected by the \ssnmt\ filtering process, we integrate the following UMT synthetic data creation techniques \emph{on-line} (Figure \ref{fig:architecture}):

\paragraph{Back-translation (BT):} Given a rejected sentence $s_{L1}$, we use the current state of the \ssnmt\ system to back-translate it into $s_{L2}^{BT}$.  The synthetic pair in the opposite direction $s_{L2}^{BT} \rightarrow s_{L1}$ is added to the batch for further training. 
We perform the same filtering process as for SPE so that only good quality back-translations are added.
We apply the same to source sentences in $L2$.

\paragraph{Word-translation (WT):} 
For synthetic sentence pairs rejected by BT filtering, we perform word-by-word translation. Given a rejected sentence $s_{L1}$ with tokens $w_{L1} \in L1$, we replace each token with its nearest neighbor $w_{L2} \in L2$ in the bilingual word embedding layer of the model to obtain $s_{L2}^{\small WT}$. 
We then train on the synthetic pair in the opposite direction $s_{L2}^{WT} \rightarrow s_{L1}$.
As with BT, this is applied to both language directions. To ensure sufficient volume of synthetic data (Figure \ref{f:enfr_experiments}, right), WT data is trained on without filtering.

\paragraph{Noise (N):} To increase robustness and variance in the training data, we add 
noise, i.e. token deletion, substitution and permutation, to copies of source sentences \citep{edunov-etal-2018-understanding} in parallel pairs identified via SPE, back-translations and word-translated sentences and, as with WT, we use these without additional filtering.

\paragraph{Initialization:} When languages are related and large amounts of training data is available, the initialization of \ssnmt\ is not important. However, similarly to UMT, initialization becomes crucial in the low-resource setting \citep{edman-etal-2020-low}. We explore four different initialization techniques: $i)$ no initialization (\emph{none}), i.e. random initialization for all model parameters, $ii)$ initialization of tied source and target side word embedding layers only via pre-trained cross-lingual word-embeddings (WE) while randomly initializing all other layers and $iii)$ initialization of all layers via denoising autoencoding (DAE) in a bilingual and $iv)$ multilingual (MDAE) setting. 

\paragraph{Finetuning (F):} When using MDAE initialization only, the following \ssnmt\ is multilingual, otherwise it is bilingual. Due to the multilingual nature of the \ssnmt\ with MDAE initialization, the performance of the individual languages can be limited by the \emph{curse of multilinguality} \citep{conneau-etal-2020-unsupervised}, where multilingual training leads to improvements on low-resource languages up to a certain point after which it decays. To alleviate this, we finetune converged multilingual \ssnmt\ models bilingually on a given language pair $L1$--$L2$.

\begin{table}[t]
\small
\centering
\begin{tabular}{l r rr rr r| r rr}
\toprule
      & \multicolumn{6}{c}{Comparable} & \multicolumn{3}{c}{Monolingual} \\
      \cmidrule(l){2-7}
      \cmidrule(l){8-10}
      & \# Art ($k$) &  VO (\%) &\multicolumn{2}{c}{\# Sent ($k$)}  & \multicolumn{2}{c}{\# Tok ($k$)} & \multicolumn{1}{c}{\# Sent ($k$)}  & \multicolumn{2}{c}{\# Tok ($k$)} \\ 
      \en--$L$ & & & \multicolumn{1}{c}{\en} & \multicolumn{1}{c}{$L$} & \multicolumn{1}{c}{\en} & \multicolumn{1}{c}{$L$} & \en/$L$ & \multicolumn{1}{c}{\en} & \multicolumn{1}{c}{$L$} \\
      \midrule
      \en--\af & 73 & 7.1 & 4,589 & 780 & 189,990 & 27,640  & 1,034 & 34,759 & 31,858 \\
      \en--\kn & 18 & 1.4 & 1,739 & 764 & 95,481 & 30,003 & 1,058 & 47,136 & 35,534  \\
      \en--\my & 19 & 2.1 & 1,505 & 477 & 82,537 & 15,313 & 997 & 43,752 & 24,094 \\
      \en--\nep & 20 & 0.6& 1,526 & 207 & 83,524 & 7,518  & 296 & 13,149 & 9,229 \\
      \en--\sw & 34 & 6.5 & 2,375 & 244 & 122,593 & 8,774  & 329 & 13,957 & 9,937 \\
      \en--\yo & 19 & 5.7 & 1,314 & 34 & 82,674 & 1,536  & 547 & 17,953 & 19,370 \\
      
\bottomrule
\end{tabular}
\caption{Number of sentences (Sent) and tokens (Tok) in the comparable and monolingual datasets. For comparable datasets, we report the number of articles (Art) and percentage of vocabulary overlap (VO) between the two languages in a pair. \# Sent of monolingual data (\en/$L$) is the same for \en\ and its corresponding $L$ due to downsampling of \en\ to match $L$. }
\label{t:corpora}
\end{table}

\section{Experimental Setting}
\label{s:setting}

\subsection{Data}
\paragraph{MT Training}

For training, we use Wikipedia (WP) as a comparable corpus and download the dumps\footnote{Dumps were downloaded on February 2021 from \url{dumps.wikimedia.org/}} and extract comparable articles per language pair (\emph{Comparable} in Table~\ref{t:corpora}) using WikiExtractor\footnote{\url{github.com/attardi/wikiextractor}}. 
For validation and testing, we use the test and development data from \citet{mckellar2020dataset}
(\en--\af),
WAT2021%
\footnote{\url{lotus.kuee.kyoto-u.ac.jp/WAT/indic-multilingual/index.html}} (\en--\kn),
WAT2020
(\en--\my) \citep{yi2018myanmar}, FLoRes
(\en--\nep) \citep{guzman-etal-2019-flores}, \citet{surafel2020low}
(\en--\sw), and MENYO-20k
(\en--\yo) \citep{adelani2021menyo20k}. 
For \en--\fr\ we use \emph{newstest2012} for development and \emph{newstest2014} for testing.
As the \en--\af\ data does not have a development split, we additionally sample 1\,$k$ sentences from CCAligned
\citep{elkishky_ccaligned_2020} to use as \en--\af\ development data. The \en--\sw\ test set is divided into several sub-domains, and we only evaluate on the TED talks domain, since the other domains are noisy, e.g. localization or religious corpora.

\paragraph{MT Initialization}
We use the monolingual Wikipedias to initialize \ssnmt. As the monolingual Wikipedia for \yoruba\ is especially small (65\,$k$ sentences), we use the \yoruba\ side of JW300 \citep{agic-vulic-2019-jw300} as additional monolingual initialization data. For each monolingual data pair \en--\{\af,...,\yo\}, the large English monolingual corpus is downsampled to its low(er)-resource counterpart before using the data (\emph{Monolingual} in Table~\ref{t:corpora}).

For the word-embedding-based initialization, we learn CBOW word embeddings using \texttt{word2vec} \citep{mikolov2013distributed},
which are then projected into a common multilingual space via \texttt{vecmap}
\citep{artetxe2017acl} to attain bilingual embeddings between $en$--\{$af$,...,$yo$\}.
For the weak-supervision of the bilingual mapping process, we use a list of numbers (\en--\fr\ only) which is augmented with 200 Swadesh list\footnote{\url{https://en.wiktionary.org/wiki/Appendix:Swadesh_lists}} entries for the low-resource experiments.

For DAE initialization, we do not use external, highly-multilingual pre-trained language models, since in practical terms these may not cover the language combination of interest\footnote{This is the case here: \texttt{MBart-50} \citep{tang2020multilingual} does not cover Kannada, Swahili and \yoruba.}. 
We therefore use the monolingual data to train a bilingual (\en+\{\af,...\yo\}) DAE using BART-style noise \citep{liu2020mbart}. We set aside 5\,$k$ sentences for testing and development each. We use BART-style noise ($\lambda=3.5$, $p=0.35$) for word sequence masking. We add one random mask insertion per sequence and perform a sequence permutation. For the multilingual DAE (MDAE) setting, we train a single denoising autoencoder on the monolingual data of all languages, where \en\ is downsampled to match the largest non-English monolingual dataset (\kn).

In all cases \ssnmt\ training is bidirectional between two languages \en--\{\af,...,\yo\}, except for MDAE, where \ssnmt\ is trained multilingually between all language combinations in \{\af,\en,...,\yo\}.

\subsection{Preprocessing}

On the Wikipedia corpora, we perform sentence tokenization using NLTK \citep{bird-2006-nltk}. For languages using Latin scripts (\af,\en,\sw,\yo) we perform punctuation normalization and truecasing using standard Moses \citep{koehn-etal-2007-moses} scripts on all datasets. For \yoruba\ only, we follow \citet{adelani2021} and perform automatic diacritic restoration.
Lastly, we perform language identification on all Wikipedia corpora using \texttt{polyglot}.\footnote{\url{https://github.com/aboSamoor/polyglot}} After exploring different byte-pair encoding (BPE) \citep{sennrich-etal-2016-neural} vocabulary sizes of 2\,$k$, 4\,$k$, 8\,$k$, 16\,$k$ and 32\,$k$, we choose 2\,$k$ (\en--\yo), 4\,$k$ (\en--\{\kn,\my,\nep,\sw\}) and 16\,$k$ (\en--\af) merge operations using \texttt{sentence-piece}\footnote{\url{https://github.com/google/sentencepiece}} \citep{kudo-richardson-2018-sentencepiece}.
We prepend a source and a target language token to each sentence.
For the \en--\fr\ experiments only, we use the data processing by \citet{ruiter2020selfinduced} in order to minimize experimental differences for later comparison.

\subsection{Model Specifications and Evaluation}
Systems are either not initialized, initialized via bilingual word embeddings, or via pre-training using (M)DAE.
Our implementation of \ssnmt\
is a 
transformer base with default parameters.
We use a batch size of 50 sentences and a maximum sequence length of 100 tokens.
For evaluation, we use BLEU \citep{papineni2002BLEU} calculated using \texttt{SacreBLEU}\footnote{\url{https://github.com/mjpost/sacrebleu}}$^{,}$\footnote{\texttt{BLEU+case.mixed+numrefs.4+smooth.exp+tok.intl+version.1.4.9}} \citep{post-2018-call}
and all confidence intervals ($p=95\%$) are calculated using bootstrap resampling~\citep{koehn-2004-statistical} as implemented in \texttt{multeval}\footnote{\url{https://github.com/jhclark/multeval}} \citep{clark-etal-2011-better}.

\section{Exploration of Corpus Sizes (\en--\fr)}
\label{s:data_sizes}

To explore which technique works best with varying data sizes, and to compare with the high-resource \ssnmt\ setting in \citet{ruiter2020selfinduced}, we train \ssnmt\ on \en--\fr, with different combinations of techniques (+BT, +WT, +N) over decreasingly small corpus sizes. The base (B) model is a simple \ssnmt\ model with SPE.

\begin{figure}
    \centering
    \includegraphics[width=0.46\columnwidth]{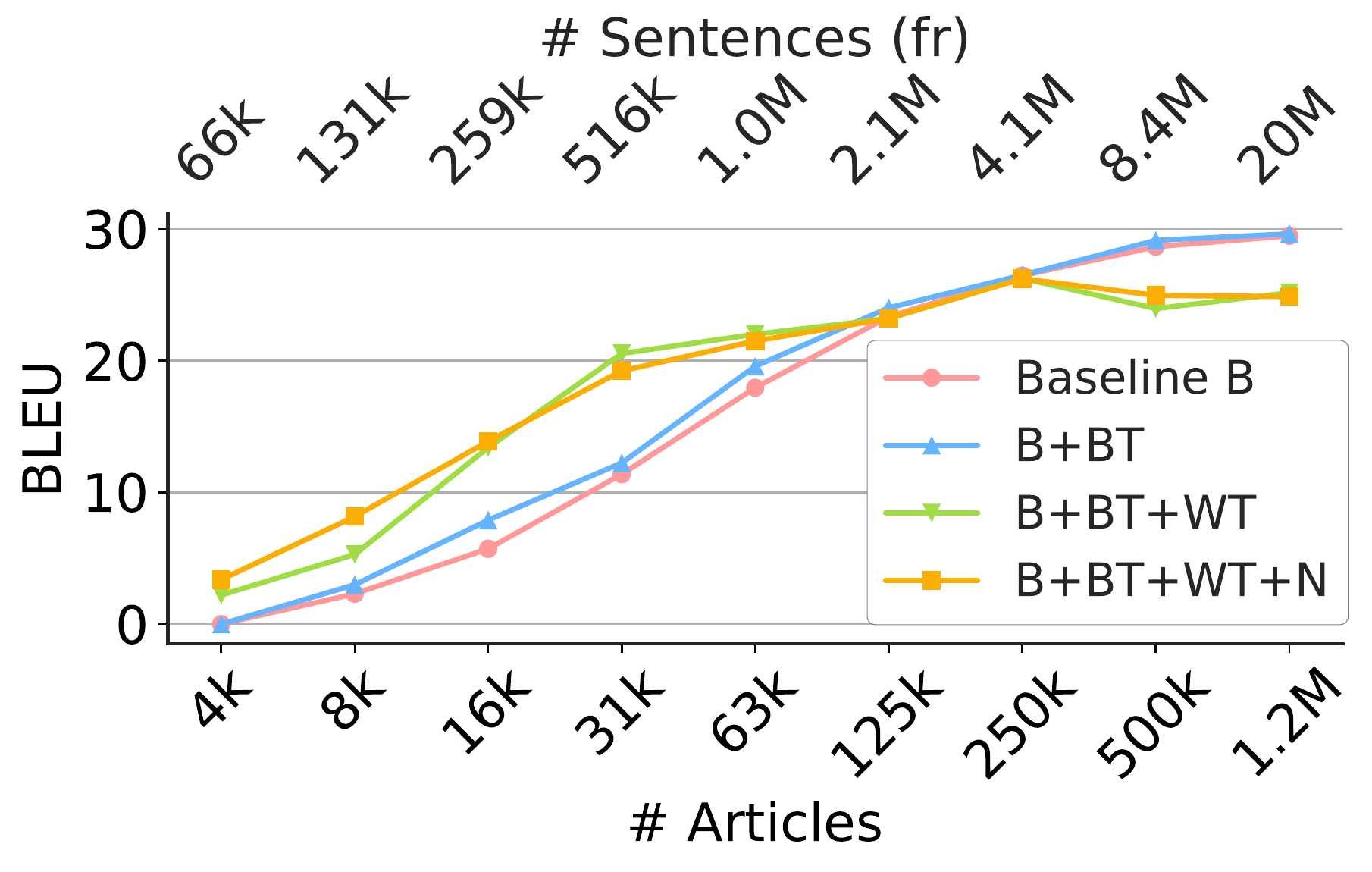}
    \hspace{7mm}
    \includegraphics[trim=0 -5mm 0 0, clip,width=0.46\columnwidth]{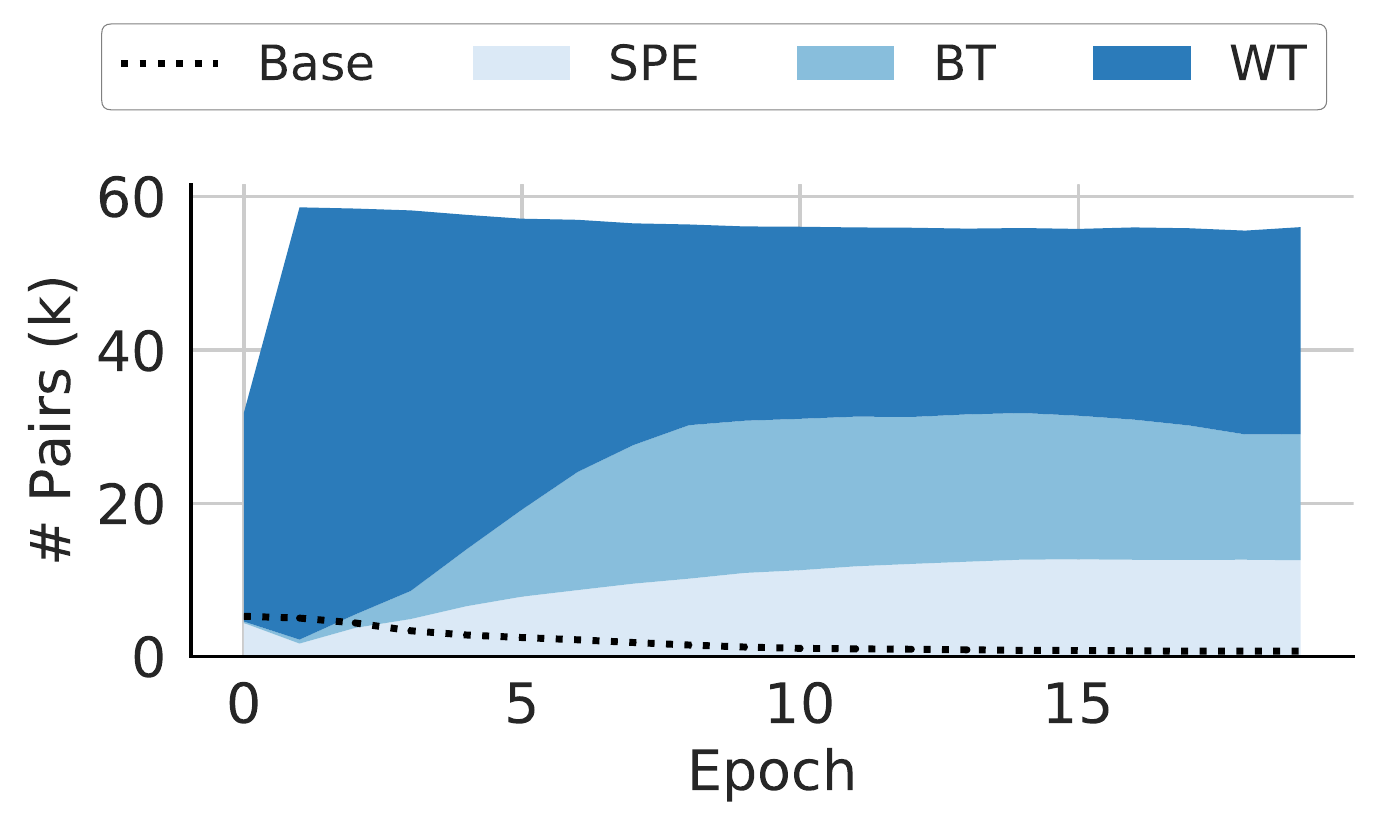}
    \caption{\textbf{Left}: BLEU scores (\en2\fr) of different techniques (+BT,+WT,+N) added to the base (B) \ssnmt\ model when trained on increasingly large numbers \en--\fr\ WP articles (\# Articles). \\
    \textbf{Right}: Number of extracted (SPE) or generated (BT,WT) sentence pairs ($k$) per technique of the B+BT+WT model trained on 4\,$k$ comparable WP articles. Number of extracted sentence pairs by the base model is shown for comparison as a dotted line.}
    \label{f:enfr_experiments}
\end{figure}

Figure \ref{f:enfr_experiments} (left) shows that translation quality as measured by BLEU is very low in the low-resource setting. For experiments with only 4\,$k$ comparable articles (similar to the corpus size available for \en--\yo),  BLEU is close to zero with base (B) and B+BT models.
Only when WT is applied to rejected back-translated pairs does training become possible, and is further improved by adding noise, yielding BLEUs of 3.38\footnote{Note that such low BLEU scores should be taken with a grain of salt: While there is an automatically measurable improvement in translation quality, a human judge would not see a meaningful improvement between different systems with low BLEU scores.} (\en2\fr) and 3.58 (\fr2\en). The maximum gain in performance obtained by WT is at 31\,$k$ comparable articles, where it adds $\sim9$ BLEU over the B+BT performance. While the additional supervisory signal provided by WT is useful in the low and medium resource settings, up until $\sim125\,k$ articles, its benefits are overcome by the noise it introduces in the high-resource scenario, leading to a drop in translation quality. Similarly, the utility of adding noise varies with corpus size. Only BT constantly adds a slight gain in performance of $\sim$1--2 over all base models, where training is possible. In the high resource case, the difference between B and B+BT is not significant, with BLEU $29.64$ (\en2\fr) and $28.56$ (\fr2\en) for B+BT, which also leads to a small, yet statistically insignificant gain over the \en--\fr\ \ssnmt\ model in \citet{ruiter2020selfinduced}, i.e. $+0.1$ (\en2\fr) and $+0.9$ (\fr2\en) BLEU.

At the beginning of training, the number of \textbf{extracted sentence pairs} (SPE) of the B+BT+WT+N model trained on the most extreme low-resource setting (4\,$k$ articles), is low (Figure \ref{f:enfr_experiments}, right), with 4\,$k$ sentence pairs extracted in the first epoch. This number drops further to 2\,$k$ extracted pairs in the second epoch, but then continually rises up to 13\,$k$ extracted pairs in the final epoch. This is not the case for the base (B) model, which starts with a similar amount of extracted parallel data but then continually extracts less as training progresses. The difference between the two models is due to the added BT and WT techniques. At the beginning of training B+BT+WT is not able to generate backtranslations of decent quality, with only few (196) backtranslations accepted for training. Rejected backtranslations are passed into WT, which leads to large numbers of WT sentence pairs up to the second epoch (56\,$k$). These make all the difference: through WT, the system is able to gain noisy supervisory signals from the data, which leads to the internal representations to become more informative for SPE, thus leading to more and better extractions. 
Then, BT and SPE enhance each other, as SPE ensures original (clean) parallel sentences to be extracted, which improves translation accuracy, and hence more and better backtranslations (e.g. up to 20\,$k$ around epoch 15) are accepted.

\section{Exploration of Language Distance}
\label{s:language_distance}

\begin{table}[t]
\centering
\footnotesize
\begin{tabular}{l ccccccc}
\toprule
           & {\bf English} & {\bf Afrikaans} & {\bf Nepali} & {\bf Kannada} & {\bf \yoruba} & {\bf Swahili} & {\bf Burmese}\\
\midrule
{\bf Typology}   & fusional$^9$ & fusional$^9$ & fusional & agglutinative & analytic & agglutinative & analytic \\
{\bf Word Order} & SVO & SOV,SVO & SOV & SOV & SOV,SVO & SVO & SOV\\
{\bf Script}     & Latin & Latin & Brahmic & Brahmic & Latin & Latin & Brahmic\\
\midrule
{\bf sim$(L$--\en)} & 1.000 & 0.822 & 0.605 & 0.602 & 0.599 & 0.456 & 0.419\\
\bottomrule
\end{tabular}
\caption{Classification (typology, word order, script) of the languages $L$ together with their cosine similarity (sim) to English based on lexical and syntactic URIEL features.}
\label{t:typology}
\end{table}

BT, WT and N data augmentation techniques are especially useful for the low- and mid-resource settings of related language pairs such as English and French (both \emph{Indo-European}). To apply the approach to truly low-resource language pairs, and to verify which language-specific characteristics impact the effectiveness of the different augmentation techniques, we train and test our model on a selected number of languages (Table \ref{t:typology}) based on their typological and graphemic distance from English (\emph{fusional}\textrightarrow\emph{analytic}\footnote{English and Afrikaans are traditionally categorized as fusional languages. However, due to their small morpheme-word ratio, both English and Afrikaans are nowadays often categorized as analytic languages.}, SVO, Latin script). Focusing on similarities on the lexical and syntactic level,\footnote{This corresponds to \texttt{lang2vec} features \texttt{syntax\_average} and \texttt{inventory\_average}.} we retrieve the URIEL \citep{littell-etal-2017-uriel} representations of the languages using \texttt{lang2vec}\footnote{\url{https://pypi.org/project/lang2vec/}} and calculate their cosine similarity to English. Afrikaans is the most similar language to English, with a similarity of $0.822$, and pre-BPE vocabulary (token) overlap of 7.1\% (Table \ref{t:corpora}), which is due to its similar typology (\emph{fusional}\textrightarrow\emph{analytic}) and comparatively large vocabulary overlap (both languages belong to the West-Germanic language branch). The most distant language is Burmese (sim $0.419$, vocabulary overlap 2.1\%), which belongs to the Sino-Tibetan language family and uses its own (Brahmic) script.

We train \ssnmt\ with combinations of BT, WT, N on the language combinations \en--\{\af,\kn,\my,\nep,\sw,\yo\} using the four different types of model initialization (none, WE, DAE, MDAE).

\begin{figure}
    \centering
    \includegraphics[width=\columnwidth]{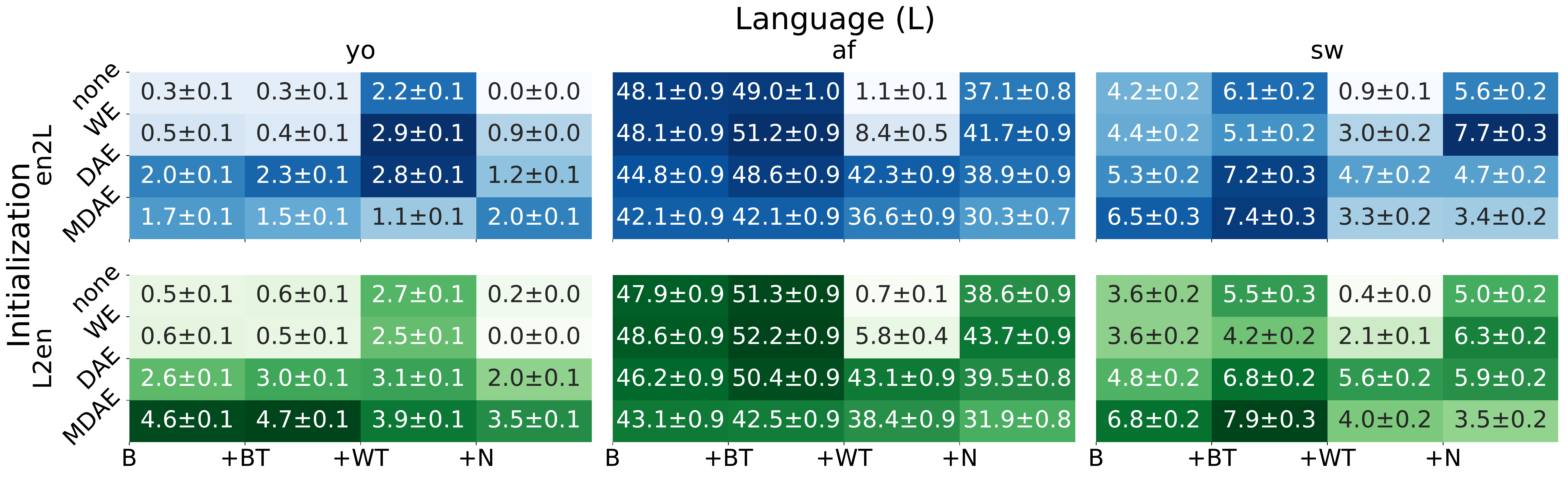}
    \includegraphics[width=\columnwidth]{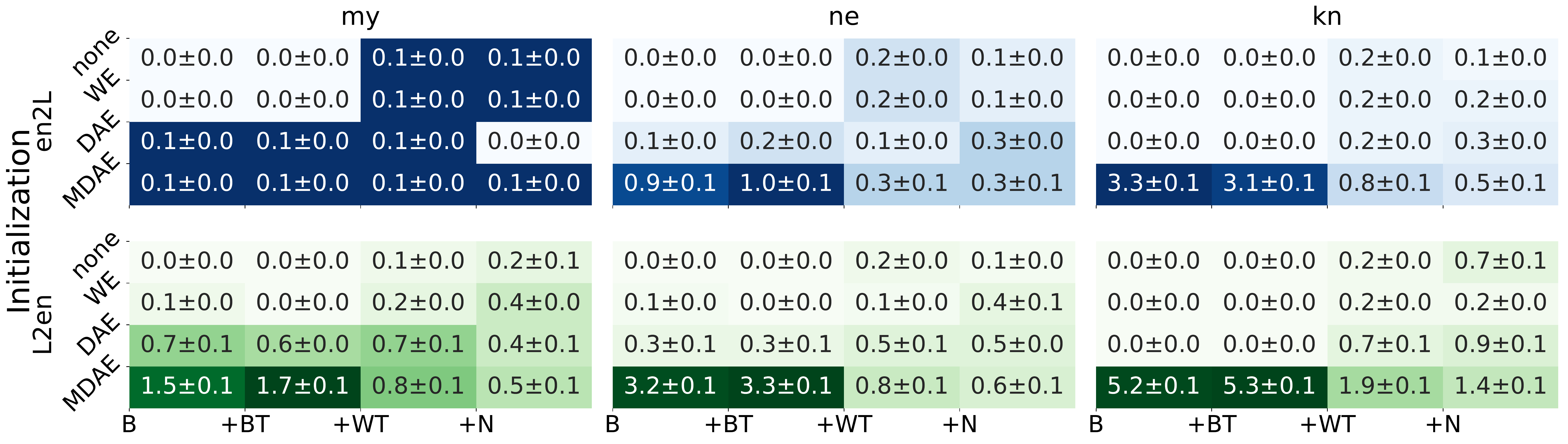}
    \caption{BLEU scores of \ssnmt\ Base (B) with added techniques (+BT,+WT,+N) on low-resource language combinations \en2$L$ and $L$2\en, with $L=\{af,kn,my,ne,sw,yo\}$.}
    \label{f:results_multi}
\end{figure}

\paragraph{Intrinsic Parameter Analysis}
We focus on the intrinsic \emph{initialization} and \emph{data augmentation technique} parameters.
The difference between no (\emph{none}) and word-embedding (\emph{WE}) \textbf{initialization} is barely significant across all language pairs and techniques (Figure \ref{f:results_multi}).
For all language pairs, except \en--\af, MDAE initialization tends to be the best choice, with major gains of $+4.2$ BLEU (\yo2\en, B+BT) and $+5.3$ BLEU (\kn2\en, B+BT) over their WE-initialized counterparts. This is natural, since pre-training on (M)DAE allows the \ssnmt\ model to learn how to generate fluent sentences. By performing (M)DAE, the model also learns to denoise noisy inputs, resulting in a big improvement in translation performance (e.g. $+37.3$ BLEU, \af2\en\ DAE) on the \en--\af\ and \en--\sw\ B+BT+WT models in comparison to their WE-initialized counterparts. Without (M)DAE pre-training, the noisy word-translations lead to very low BLEU scores. Adding an additional denoising task, either via (M)DAE initialization or via adding the +N data augmentation technique, lets the model also learn from noisy word-translations with improved results. For \en--\af\ only, the WE initialization generally performs best, with BLEU scores of $52.2$ (\af2\en) and $51.2$ (\en2\af).
For language pairs using different scripts, i.e. Latin--Brahmic (\en--\{\kn,\my,\nep\}), the gain by performing bilingual DAE pre-training is negligible, as results are generally low. These languages also have a different word order (SOV) than English (SVO), which may further increase the difficulty of the translation task \citep{banerjee2019ordering,kim-etal-2020-unsupervised}. However, once the pre-training and MT learning is multilingual (MDAE), the different language directions benefit from another and an internal mapping of the languages into a shared space is achieved. This leads to BLEU scores of $1.7$ (\my2\en), $3.3$ (\nep2\en) and $5.3$ (\kn2\en) using the B+BT technique. The method is also beneficial when translating into the low-resource languages, with \en2\kn\ reaching BLEU $3.3$ (B).

\begin{figure}
    \centering
    \includegraphics[width=\columnwidth]{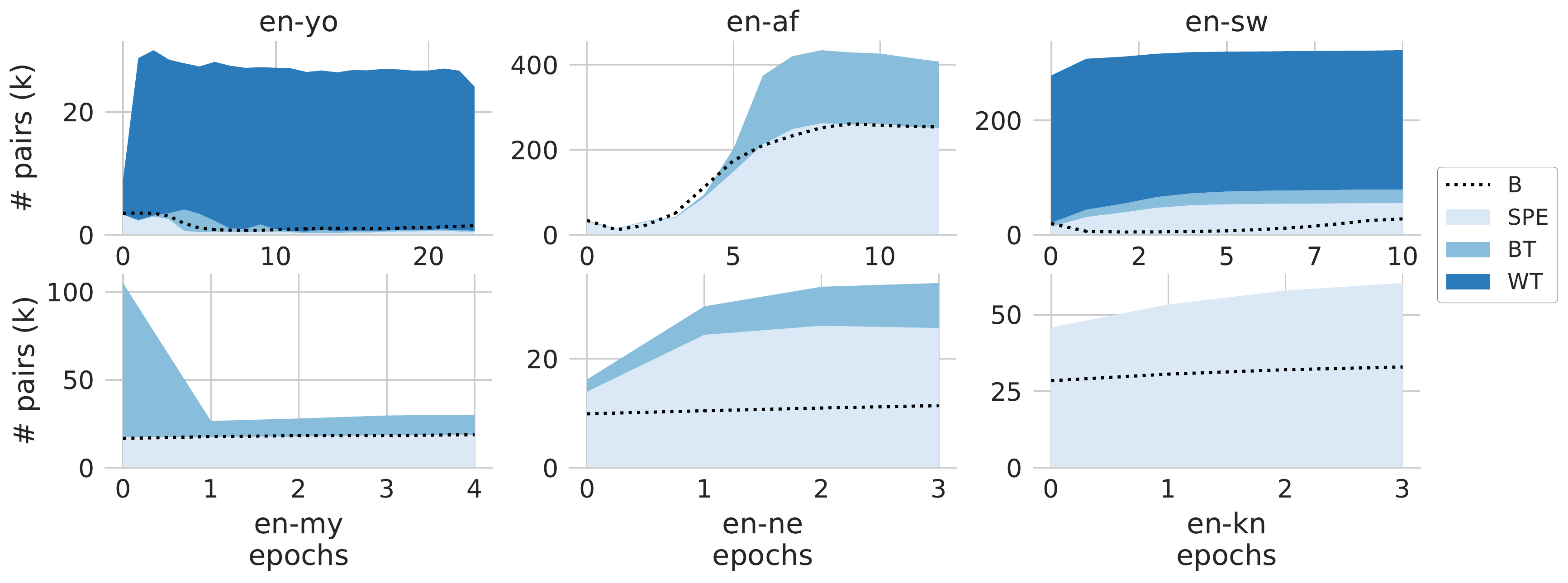}
    \caption{Number of extracted (SPE) or generated (BT,WT) sentence pairs ($k$) per technique of the best performing \ssnmt\ model (\en2$L$) per language $L$. Number of extracted sentence pairs by the base model (B) are shown for comparison as a dotted line.}
    \label{f:technique_lr}
\end{figure}

B+BT+WT seems to be the best \textbf{data augmentation technique} when the amount of data is very small, as is the case for \en--\yo, with gains of $+2.4$ BLEU on \en2\yo\ over the baseline B. This underlines the findings in Section \ref{s:data_sizes}, that WT serves as a crutch to start the extraction and training of \ssnmt. Further adding noise (+N) tends to adversely impact on results on this language pair. On languages with more data available (\en--\{\af,\kn,\my,\nep,\sw\}), +BT tends to be the best choice, with top BLEUs on \en--\sw\ of $7.4$ (\en2\sw, MDAE) and $7.9$ (\sw2\en, MDAE).  This is  due to these models being able to sufficiently learn on B (+BT) only (Figure \ref{f:technique_lr}), thus not needing +WT as a crutch to start the extraction and MT learning process. Adding +WT to the system only adds additional noise and thus makes results worse.

\paragraph{Extrinsic Parameter Analysis}
We focus on the extrinsic parameters \emph{linguistic distance} and \emph{data size}. 
Our model is able to learn MT also on \textbf{distant language pairs} such as \en--\sw\ (sim $0.456$), with top BLEUs of $7.7$ (\en2\sw, B+BT+W+N) and $7.9$ (\sw2\en, B+BT). Despite being typologically closer, training \ssnmt\ on \en--\nep\ (sim $0.605$) only yields BLEUs above 1 in the multilingual setting (BLEU $3.3$ \nep2\en). This is the case for all languages using a different script than English (\kn,\my,\nep), underlining the fact that achieving a cross-lingual representation, i.e. via multilingual (pre-)training or a decent overlap in the (BPE) vocabulary (as in \en--\{\af,\sw,\yo\}) of the two languages, is vital for identifying good similar sentence pairs at the beginning of training and thus makes training possible. For \en--\my\, the MDAE approach was only beneficial in the \my2\en\ direction, but had no effect on \en2\my, which may be due to the fact that \my\ is the most distant language from \en\ (sim 0.419) and, contrary to the other low-resource languages we explore, does not have any related language\footnote{Both Nepali and Kannada share influences from Sanskrit. Swahili and \yoruba\ are both Niger-Congo languages, while English and Afrikaans are both Indo-European.} in our experimental setup, which makes it difficult to leverage supervisory signals from a related language.

When the \textbf{amount of data} is small (\en--\yo), the model does not achieve BLEUs above $1$ without the WT technique or without (M)DAE initialization, since the extraction recall of a simple \ssnmt\ system is low at the beginning of training \citep{ruiter2020selfinduced} and thus SPE fails to identify sufficient parallel sentences to improve the internal representations, which would then improve SPE recall. This is analogous to the observations on the \en-\fr\ base model B trained on $4\,k$ WP articles (Figure \ref{f:enfr_experiments}). Interestingly, the differences between no/WE and DAE initialization are minimized when using WT as a data augmentation technique, showing that it is an effective method that makes pre-training unnecessary when only small amounts of data are available. For larger data sizes (\en--\{\af,\sw\}), the opposite is the case: the models sufficiently learn SPE and MT without WT, and thus WT just adds additional noise.

\paragraph{Extraction and Generation}

The SPE extraction and BT/WT generation curves (Figure \ref{f:technique_lr}) for \en--\af\ (B+BT, WE) are similar to those on \en--\fr\ (Figure \ref{f:enfr_experiments}, right). At the beginning of training, not many pairs (32\,$k$) are extracted, but as training progresses, the model internal representations are improved and it starts extracting more and more parallel data, up to 252\,$k$ in the last epoch. Simultaneously, translation quality improves and the number of backtranslations generated increases drastically from 2\,$k$ up to 156\,$k$ per epoch. However, as the amount of data for \en--\af\ is large, the base model B has a similar extraction curve. Nevertheless, translation quality is improved by the additional backtranslations  ($+3.1$ BLEU). For \en--\sw\ (B+BT+WT+N, WE), the curves are similar to those of \en--\fr, where the added word-translations serve as a crutch to make SPE and BT possible, thus showing a gap between the number of extracted sentences (SPE) ($\sim5.5\,k$) of the best model and those of the baseline (B) ($\sim$1--2\,$k$). For \en--\yo\ (B+BT+WT, WE), the amount of extracted data is very small ($\sim0.5\,k$) for both the baseline and the best model. Here, WT fails to serve as a crutch as the number of extractions does not increase, but instead is overwhelmed by the number of word translations. For \en--\{\kn,\nep\} (MDAE), the extraction and BT curves also rise over time. For \en--\my, where all training setups show similar translation performance in the \en2\my\ direction, we show the extraction and BT curves for B+BT with WE initialization. We observe that, as opposed to all other models, both lines are flat, underlining the fact that due to the lack of sufficiently cross-lingual model-internal representations, the model does not enter the self-supervisory cycle common to \ssnmt.

\paragraph{Bilingual Finetuning}

\begin{table}[t]
\centering
\small
\begin{tabular}{l cccccccccccc}
\toprule
           & \multicolumn{2}{c}{\bf \en--\af} & \multicolumn{2}{c}{\bf \en--\kn} & \multicolumn{2}{c}{\bf \en--\my} & \multicolumn{2}{c}{\bf \en--\nep} & \multicolumn{2}{c}{\bf \en--\sw} & \multicolumn{2}{c}{\bf \en--\yo}\\
          & \textrightarrow & \textleftarrow & \textrightarrow & \textleftarrow & \textrightarrow & \textleftarrow & \textrightarrow & \textleftarrow & \textrightarrow & \textleftarrow & \textrightarrow & \textleftarrow\\
\midrule
Best* &          \textbf{51.2} & \textbf{52.2} & 0.3 & 0.9 & 0.1 & 0.7 & 0.3 & 0.5 & 7.7 & 6.8 & \textbf{2.9} & 3.1 \\
MDAE &     42.5 & 42.5 & 3.1 & 5.3 & 0.1 & 1.7 & 1.0 & 3.3 & 7.4 & 7.9 & 1.5 & 4.7 \\ \midrule
MDAE+F &  46.3 & 50.2 & \textbf{5.0} & \textbf{9.0} & \textbf{0.2} & \textbf{2.8} & \textbf{2.3} & \textbf{5.7} & \textbf{11.6} & \textbf{11.2} & \textbf{2.9} & \textbf{5.8} \\
\bottomrule
\end{tabular}
\caption{BLEU scores on the \en2$L$ (\textrightarrow) and $L$2\en\ (\textleftarrow) directions of top performing \ssnmt\ model without finetuning and without MDAE (Best*) and \ssnmt\ using MDAE initialization and B+BT technique with (MDAE+F) and without finetuning (MDAE).}
\label{t:finetuning}
\end{table}

The overall trend shows that MDAE pre-training with multilingual \ssnmt\ training in combination with back-translation (B+BT) leads to top results for low-resource similar and distant language combinations. For \en--\af only, which has more comparable data available for training and is a very similar language pair, the multilingual setup is less beneficial. The model attains enough supervisory signals when training bilingually on \en--\af, thus the additional languages in the multilingual setup are simply noise for the system. While the MDAE setup with multilingual MT training makes it possible to map distant languages into a shared space and learn MT, we suspect that the final MT performance on the individual language directions is ultimately being held back due to the multilingual noise of other language combinations. To verify this, we use the converged MDAE B+BT model and fine-tune it using the B+BT approach on the different \en--\{\af,...,\yo\} combinations individually (Table \ref{t:finetuning}).

In all cases, the bilingual finetuning improves the multilingual model, with a major increase of $+4.2$ BLEU for \en--\sw\ resulting in a BLEU score of 11.6. The finetuned models almost always produce the best performing model, showing that the process of $i)$ multilingual pre-training (MDAE) to achieve a cross-lingual representation, $ii)$ \ssnmt\ online data extraction (SPE) with online back-translation (B+BT) to obtain increasing quantities of supervisory signals from the data, followed by $iii)$ focused bilingual fine-tuning to remove multilingual noise is key to learning low-resource MT also on distant languages without the need of any parallel data.

\section{Comparison to other NMT Architectures}
\label{s:external_comparison}

\begin{table}[t]
\small
\centering
\setlength{\tabcolsep}{3pt} 

\begin{tabular}{l | llr | r r r r | r r}
 \toprule
Pair       & Init. & Config.  & Best  & Base & UMT & UMT+NMT & Laser~ & ~~TSS &\#P ($k$)   \\ \midrule
\en2\af & WE & B+BT & \textbf{51.2$\pm$.9}& 48.1$\pm$.9 & 27.9$\pm$.8 & 44.2$\pm$.9 & \textbf{52.1$\pm$1.0} & 35.3 & 37 \\
\af2\en & WE & B+BT & \textbf{52.2$\pm$.9} & 47.9$\pm$.9 & 1.4$\pm$.1 & 0.7$\pm$.1 & \textbf{52.9$\pm$.9} & -- & -- \\ \midrule
\en2\kn & MDAE & B+BT+F & \textbf{5.0$\pm$.2} & 0.0$\pm$.0 & 0.0$\pm$.0 & 0.0$\pm$.0 & -- & 21.3 & 397 \\
\kn2\en & MDAE & B+BT+F & \textbf{9.0$\pm$.2} & 0.0$\pm$.0 & 0.0$\pm$.0 & 0.0$\pm$.0 & -- & 40.3 & 397 \\ \midrule
\en2\my & MDAE & B+BT+F & \textbf{0.2$\pm$.0} & 0.0$\pm$.0 & 0.1$\pm$.0 & 0.0$\pm$.0 & 0.0$\pm$.0 & 39.3 & 223 \\
\my2\en & MDAE & B+BT+F & \textbf{2.8$\pm$.1} & 0.0$\pm$.0 & 0.0$\pm$.0 & 0.0$\pm$.0 & 0.1$\pm$.0 & 38.6 & 223 \\ \midrule
\en2\nep & MDAE & B+BT+F & \textbf{2.3$\pm$.1} & 0.0$\pm$.0 & 0.1$\pm$.0 & 0.0$\pm$.0 & 0.5$\pm$.1 & 8.8 & -- \\
\nep2\en & MDAE & B+BT+F & \textbf{5.7$\pm$.2} & 0.0$\pm$.0 & 0.0$\pm$.0 & 0.0$\pm$.0 & 0.2$\pm$.0 & 21.5 & -- \\ \midrule
\en2\sw & MDAE & B+BT+F & \textbf{11.6$\pm$.3} & 4.2$\pm$.2 & 3.6$\pm$.2 & 0.2$\pm$.0 & 10.0$\pm$.3 & 14.8 & 995 \\
\sw2\en & MDAE & B+BT+F & \textbf{11.2$\pm$.3} & 3.6$\pm$.2 & 0.3$\pm$.0 & 0.0$\pm$.0 & 8.4$\pm$.3 & 19.7 & 995 \\ \midrule
\en2\yo & MDAE & B+BT+F & \textbf{2.9$\pm$.1} & 0.3$\pm$.1 & 1.0$\pm$.1 & 0.3$\pm$.1 & -- & 12.3 & 501 \\
\yo2\en & MDAE & B+BT+F & \textbf{5.8$\pm$.1} & 0.5$\pm$.1 & 0.6$\pm$.0 & 0.0$\pm$.0 & -- & 22.4 & 501 \\

\bottomrule
\end{tabular}
\caption{BLEU scores of the best \ssnmt\ configuration (columns 2-4) compared with \ssnmt\ base, USMT(+UNMT) and a supervised NMT system trained on Laser extractions (columns 5-8). Top scoring systems (TSS) per test set and the amount of parallel training sentences (\#P) available for reference (columns 9-10).
}
\label{t:bleus}
\end{table}

We compare the best \ssnmt\ model configuration per language pair with the \ssnmt\ \textbf{baseline} system, and with Monoses \citep{artetxe2019effective}, an \textbf{unsupervised} machine translation model in its statistical (USMT) and hybrid (USMT+UNMT) version (Table \ref{t:bleus}). Over all languages, \ssnmt\ with data augmentation outperforms both the \ssnmt\ baseline and UMT models.

We also compare our results with a \textbf{supervised} NMT system trained on WP parallel sentences \textbf{extracted} by Laser\footnote{\url{https://github.com/facebookresearch/LASER}} \citep{artetxe2018massively} (\en--\{\af,\my\}) in a preprocessing data extraction step with the recommended extraction threshold of $1.04$. 
We use the pre-extracted and similarity-ranked WikiMatrix~\citep{schwenk2019wikimatrix} corpus, which uses Laser to extract parallel sentences, for \en--\{\nep,\sw\}. Laser is not trained on \kn\ and \yo, thus these languages are not included in the analysis. For \en--\af, our model and the supervised model trained on Laser extractions perform equally well. In all other cases, our model statistically significantly outperforms the supervised LASER model, which is surprising, given the fact that the underlying LASER model was trained on parallel data in a highly multilingual setup (93 languages), while our MDAE setup does not use any parallel data and was trained on the monolingual data of much fewer language directions (7 languages) only. This again underlines the effectiveness of joining \ssnmt\ with BT, multilingual pre-training and bilingual finetuning.

For reference, we also report the \textbf{top-scoring system} (TSS) per language direction based on top results reported on the relevant test sets together with the amount of parallel training data available to TSS systems. In case of language pairs whose test set is part of ongoing shared tasks (\en--\{\kn,\my\}), we report the most recent results reported on the shared task web-pages (Section \ref{s:setting}). The amount of parallel data available for these TSS varies greatly across languages, from 37\,$k$ (\en--\af) to 995\,$k$  (often noisy) sentences. 
In general, TSS systems perform much better than any of the \ssnmt\ configurations or unsupervised models. This is natural, as TSS systems are mostly supervised \citep{martinus2019focus,adelani2021menyo20k}, semi-supervised \citep{surafel2020low} or multilingual models with parallel pivot language pairs \citep{guzman-etal-2019-flores}, none of which is used in the UMT and \ssnmt\ models. For \en2\af\ only, our best configuration and the supervised NMT model trained on Laser extractions outperform the current TSS, with a gain in BLEU of $+16.9$ (B+BT), which may be due to the small amount of parallel data the TSS was trained on (37\,$k$ parallel sentences).

\section{Discussion and Conclusion}
\label{s:conclusion}

Across all tested low-resource language pairs, joining \ssnmt-style online sentence pair extraction with UMT-style online back-translation significantly outperforms the \ssnmt\ baseline and unsupervised MT models, indicating that the small amount of available supervisory signals in the data is exploited more efficiently. 
Our models also outperform supervised NMT systems trained on Laser extractions, which is remarkable given that our systems are trained on non-parallel data only, while Laser has been trained on massive amounts of parallel data.

While \ssnmt\ with data augmentation and MDAE pre-training is able to learn MT even on a low-resource distant language pair such as \en--\kn, it can fail when a language does not have any relation to other languages included in the multilingual pre-training, which was the case for \my\ in our setup. This can be overcome by being conscientious of the importance of language distance and including related languages during MDAE pre-training and \ssnmt\ training.
We make our code and data publicly available.\footnote{\url{https://github.com/ruitedk6/comparableNMT}}

\section*{Acknowledgements}
We thank David Adelani and Jesujoba Alabi for their insights on \yoruba.
Part of this research was made possible through a research award from Facebook AI.
Partially funded by the German Federal Ministry of Education and Research under the funding code 01IW20010 (Cora4NLP). The authors are responsible for the content of this publication. 

\small

\bibliographystyle{apalike}
\bibliography{mtsummit2021}

\begin{thebibliography}{}

\bibitem[Adelani et~al., 2021a]{adelani2021menyo20k}
Adelani, D.~I., Ruiter, D., Alabi, J.~O., Adebonojo, D., Ayeni, A., Adeyemi,
  M., Awokoya, A., and Espa{\~{n}}a{-}Bonet, C. (2021a).
\newblock {MENYO-20k: {A} Multi-domain English-Yor{\`{u}}b{\'{a}} Corpus for
  Machine Translation and Domain Adaptation}.
\newblock {\em AfricaNLP Workshop, CoRR}, abs/2103.08647.

\bibitem[Adelani et~al., 2021b]{adelani2021}
Adelani, D.~I., Ruiter, D., Alabi, J.~O., Adebonojo, D., Ayeni, A., Adeyemi,
  M., Awokoya, A., and Espa{\~n}a-Bonet, C. (2021b).
\newblock {The Effect of Domain and Diacritics in {Y}or{\`u}b{\'a}--English
  Neural Machine Translation}.
\newblock In {\em Proceedings of Machine Translation Summit (Research Track)}.
  European Association for Machine Translation.

\bibitem[Agi{\'c} and Vuli{\'c}, 2019]{agic-vulic-2019-jw300}
Agi{\'c}, {\v{Z}}. and Vuli{\'c}, I. (2019).
\newblock {JW}300: A wide-coverage parallel corpus for low-resource languages.
\newblock In {\em Proceedings of the 57th Annual Meeting of the Association for
  Computational Linguistics}, pages 3204--3210, Florence, Italy. Association
  for Computational Linguistics.

\bibitem[Artetxe et~al., 2017]{artetxe2017acl}
Artetxe, M., Labaka, G., and Agirre, E. (2017).
\newblock Learning bilingual word embeddings with (almost) no bilingual data.
\newblock In {\em Proceedings of the 55th Annual Meeting of the Association for
  Computational Linguistics (Volume 1: Long Papers)}, pages 451--462.

\bibitem[Artetxe et~al., 2019]{artetxe2019effective}
Artetxe, M., Labaka, G., and Agirre, E. (2019).
\newblock An effective approach to unsupervised machine translation.
\newblock In {\em Proceedings of the 57th Annual Meeting of the Association for
  Computational Linguistics}, pages 194--203, Florence, Italy. Association for
  Computational Linguistics.

\bibitem[Artetxe et~al., 2018]{artetxeEtAl:ICLR:2018}
Artetxe, M., Labaka, G., Agirre, E., and Cho, K. (2018).
\newblock Unsupervised neural machine translation.
\newblock In {\em Proceedings of the Sixth International Conference on Learning
  Representations, ICLR}.

\bibitem[Artetxe and Schwenk, 2019a]{artetxe2018margin}
Artetxe, M. and Schwenk, H. (2019a).
\newblock Margin-based parallel corpus mining with multilingual sentence
  embeddings.
\newblock In {\em Proceedings of the 58th Annual Meeting of the Association for
  Computational Linguistics}, pages 3197--3203, Florence, Italy. Association
  for Computational Linguistics.

\bibitem[Artetxe and Schwenk, 2019b]{artetxe2018massively}
Artetxe, M. and Schwenk, H. (2019b).
\newblock Massively multilingual sentence embeddings for zero-shot
  cross-lingual transfer and beyond.
\newblock {\em Transactions of the Association for Computational Linguistics},
  7:597--610.

\bibitem[Banerjee et~al., 2019]{banerjee2019ordering}
Banerjee, T., Murthy, V.~R., and Bhattacharyya, P. (2019).
\newblock Ordering matters: Word ordering aware unsupervised {NMT}.
\newblock {\em CoRR}, abs/1911.01212.

\bibitem[Barrault et~al., 2020]{WMT:2020}
Barrault, L., Bojar, O., Bougares, F., Chatterjee, R., Costa-jussà, M.~R.,
  Federmann, C., Fishel, M., Fraser, A., Graham, Y., Guzman, P., Haddow, B.,
  Huck, M., Yepes, A.~J., Koehn, P., Martins, A., Morishita, M., Monz, C.,
  Nagata, M., Nakazawa, T., and Negri, M., editors (2020).
\newblock {\em Proceedings of the Fifth Conference on Machine Translation}.
\newblock Association for Computational Linguistics, Online.

\bibitem[Bird, 2006]{bird-2006-nltk}
Bird, S. (2006).
\newblock {NLTK}: The {N}atural {L}anguage {T}oolkit.
\newblock In {\em Proceedings of the {COLING}/{ACL} 2006 Interactive
  Presentation Sessions}, pages 69--72, Sydney, Australia. Association for
  Computational Linguistics.

\bibitem[Bojar and Tamchyna, 2011]{bojar-tamchyna-2011-improving}
Bojar, O. and Tamchyna, A. (2011).
\newblock Improving translation model by monolingual data.
\newblock In {\em Proceedings of the Sixth Workshop on Statistical Machine
  Translation}, pages 330--336, Edinburgh, Scotland. Association for
  Computational Linguistics.

\bibitem[Chaudhary et~al., 2019]{chaudhary-EtAl:2019:WMT}
Chaudhary, V., Tang, Y., Guzmán, F., Schwenk, H., and Koehn, P. (2019).
\newblock Low-resource corpus filtering using multilingual sentence embeddings.
\newblock In {\em Proceedings of the Fourth Conference on Machine Translation
  (Volume 3: Shared Task Papers, Day 2)}, pages 263--268, Florence, Italy.
  Association for Computational Linguistics.

\bibitem[Clark et~al., 2011]{clark-etal-2011-better}
Clark, J.~H., Dyer, C., Lavie, A., and Smith, N.~A. (2011).
\newblock Better hypothesis testing for statistical machine translation:
  Controlling for optimizer instability.
\newblock In {\em Proceedings of the 49th Annual Meeting of the Association for
  Computational Linguistics: Human Language Technologies}, pages 176--181,
  Portland, Oregon, USA. Association for Computational Linguistics.

\bibitem[Conneau et~al., 2020]{conneau-etal-2020-unsupervised}
Conneau, A., Khandelwal, K., Goyal, N., Chaudhary, V., Wenzek, G., Guzm{\'a}n,
  F., Grave, E., Ott, M., Zettlemoyer, L., and Stoyanov, V. (2020).
\newblock Unsupervised cross-lingual representation learning at scale.
\newblock In {\em Proceedings of the 58th Annual Meeting of the Association for
  Computational Linguistics}, pages 8440--8451, Online. Association for
  Computational Linguistics.

\bibitem[Conneau and Lample, 2019]{lample2019cross}
Conneau, A. and Lample, G. (2019).
\newblock Cross-lingual language model pretraining.
\newblock In Wallach, H., Larochelle, H., Beygelzimer, A., d\textquotesingle
  Alch\'{e}-Buc, F., Fox, E., and Garnett, R., editors, {\em Advances in Neural
  Information Processing Systems}, volume~32. Curran Associates, Inc.

\bibitem[Currey et~al., 2017]{currey-etal-2017-copied}
Currey, A., Miceli~Barone, A.~V., and Heafield, K. (2017).
\newblock Copied monolingual data improves low-resource neural machine
  translation.
\newblock In {\em Proceedings of the Second Conference on Machine Translation},
  pages 148--156, Copenhagen, Denmark. Association for Computational
  Linguistics.

\bibitem[Edman et~al., 2020]{edman-etal-2020-low}
Edman, L., Toral, A., and van Noord, G. (2020).
\newblock Low-resource unsupervised {NMT}: Diagnosing the problem and providing
  a linguistically motivated solution.
\newblock In {\em Proceedings of the 22nd Annual Conference of the European
  Association for Machine Translation}, pages 81--90, Lisboa, Portugal.
  European Association for Machine Translation.

\bibitem[Edunov et~al., 2018]{edunov-etal-2018-understanding}
Edunov, S., Ott, M., Auli, M., and Grangier, D. (2018).
\newblock Understanding back-translation at scale.
\newblock In {\em Proceedings of the 2018 Conference on Empirical Methods in
  Natural Language Processing}, pages 489--500, Brussels, Belgium. Association
  for Computational Linguistics.

\bibitem[El-Kishky et~al., 2020]{elkishky_ccaligned_2020}
El-Kishky, A., Chaudhary, V., Guzmán, F., and Koehn, P. (2020).
\newblock {CCAligned}: A massive collection of cross-lingual web-document
  pairs.
\newblock In {\em Proceedings of the 2020 Conference on Empirical Methods in
  Natural Language Processing (EMNLP 2020)}, pages 5960--5969, Online.
  Association for Computational Linguistics.

\bibitem[Espa{\~{n}}a{-}Bonet et~al., 2019]{espanaEtAl:WMT:2019}
Espa{\~{n}}a{-}Bonet, C., Ruiter, D., and van Genabith, J. (2019).
\newblock {UdS-DFKI Participation at WMT 2019: Low-Resource ($en$--$gu$) and
  Coreference-Aware ($en$--$de$) Systems}.
\newblock In {\em Proceedings of the Fourth Conference on Machine Translation},
  pages 382--389, Florence, Italy. Association for Computational Linguistics.

\bibitem[Gulcehre et~al., 2015]{gulcehre2015using}
Gulcehre, C., Firat, O., Xu, K., Cho, K., Barrault, L., Lin, H.-C., Bougares,
  F., Schwenk, H., and Bengio, Y. (2015).
\newblock On using monolingual corpora in neural machine translation.
\newblock {\em arXiv preprint arXiv:1503.03535}.

\bibitem[Guzm{\'a}n et~al., 2019]{guzman-etal-2019-flores}
Guzm{\'a}n, F., Chen, P.-J., Ott, M., Pino, J., Lample, G., Koehn, P.,
  Chaudhary, V., and Ranzato, M. (2019).
\newblock The {FLORES} evaluation datasets for low-resource machine
  translation: {N}epali{--}{E}nglish and {S}inhala{--}{E}nglish.
\newblock In {\em Proceedings of the 2019 Conference on Empirical Methods in
  Natural Language Processing and the 9th International Joint Conference on
  Natural Language Processing (EMNLP-IJCNLP)}, pages 6098--6111, Hong Kong,
  China. Association for Computational Linguistics.

\bibitem[Hoang et~al., 2018]{hoang-etal-2018-iterative}
Hoang, V. C.~D., Koehn, P., Haffari, G., and Cohn, T. (2018).
\newblock Iterative back-translation for neural machine translation.
\newblock In {\em Proceedings of the 2nd Workshop on Neural Machine Translation
  and Generation}, pages 18--24, Melbourne, Australia. Association for
  Computational Linguistics.

\bibitem[Johnson et~al., 2017]{johnson2016google}
Johnson, M., Schuster, M., Le, Q.~V., Krikun, M., Wu, Y., Chen, Z., Thorat, N.,
  Vi{\'{e}}gas, F.~B., Wattenberg, M., Corrado, G., Hughes, M., and Dean, J.
  (2017).
\newblock {Google's Multilingual Neural Machine Translation System: Enabling
  Zero-Shot Translation}.
\newblock {\em Transactions of the Association for Computational Linguist},
  5:339--351.

\bibitem[Karakanta et~al., 2018]{Karakanta2018}
Karakanta, A., Dehdari, J., and van Genabith, J. (2018).
\newblock Neural machine translation for low-resource languages without
  parallel corpora.
\newblock {\em Machine Translation}, 32(1):167--189.

\bibitem[Kim et~al., 2020]{kim-etal-2020-unsupervised}
Kim, Y., Gra{\c{c}}a, M., and Ney, H. (2020).
\newblock When and why is unsupervised neural machine translation useless?
\newblock In {\em Proceedings of the 22nd Annual Conference of the European
  Association for Machine Translation}, pages 35--44, Lisboa, Portugal.
  European Association for Machine Translation.

\bibitem[Koehn, 2004]{koehn-2004-statistical}
Koehn, P. (2004).
\newblock Statistical significance tests for machine translation evaluation.
\newblock In {\em Proceedings of the 2004 Conference on Empirical Methods in
  Natural Language Processing}, pages 388--395, Barcelona, Spain. Association
  for Computational Linguistics.

\bibitem[Koehn et~al., 2007]{koehn-etal-2007-moses}
Koehn, P., Hoang, H., Birch, A., Callison-Burch, C., Federico, M., Bertoldi,
  N., Cowan, B., Shen, W., Moran, C., Zens, R., Dyer, C., Bojar, O.,
  Constantin, A., and Herbst, E. (2007).
\newblock {M}oses: Open source toolkit for statistical machine translation.
\newblock In {\em Proceedings of the 45th Annual Meeting of the Association for
  Computational Linguistics Companion Volume Proceedings of the Demo and Poster
  Sessions}, pages 177--180, Prague, Czech Republic. Association for
  Computational Linguistics.

\bibitem[Koneru et~al., 2021]{koneru-etal-2021-unsupervised}
Koneru, S., Liu, D., and Niehues, J. (2021).
\newblock Unsupervised machine translation on {D}ravidian languages.
\newblock In {\em Proceedings of the First Workshop on Speech and Language
  Technologies for Dravidian Languages}, pages 55--64, Kyiv. Association for
  Computational Linguistics.

\bibitem[Kudo and Richardson, 2018]{kudo-richardson-2018-sentencepiece}
Kudo, T. and Richardson, J. (2018).
\newblock {S}entence{P}iece: A simple and language independent subword
  tokenizer and detokenizer for neural text processing.
\newblock In {\em Proceedings of the 2018 Conference on Empirical Methods in
  Natural Language Processing: System Demonstrations}, pages 66--71, Brussels,
  Belgium. Association for Computational Linguistics.

\bibitem[Kuwanto et~al., 2021]{kuwanto2021low}
Kuwanto, G., Aky{\"{u}}rek, A.~F., Tourni, I.~C., Li, S., and Wijaya, D.
  (2021).
\newblock Low-resource machine translation for low-resource languages:
  Leveraging comparable data, code-switching and compute resources.
\newblock {\em CoRR}, abs/2103.13272.

\bibitem[Lakew et~al., 2021]{surafel2020low}
Lakew, S.~M., Negri, M., and Turchi, M. (2021).
\newblock {Low Resource Neural Machine Translation: {A} Benchmark for Five
  African Languages}.
\newblock {\em AfricaNLP Workshop, CoRR}, abs/2003.14402.

\bibitem[Lample et~al., 2018a]{lampleEtAl:ICLR:2018}
Lample, G., Conneau, A., Denoyer, L., and Ranzato, M. (2018a).
\newblock Unsupervised machine translation using monolingual corpora only.
\newblock In {\em Proceedings of the Sixth International Conference on Learning
  Representations, ICLR}.

\bibitem[Lample et~al., 2018b]{lampleEtAl:EMNLP:2018}
Lample, G., Ott, M., Conneau, A., Denoyer, L., and Ranzato, M. (2018b).
\newblock Phrase-based {\&} neural unsupervised machine translation.
\newblock In {\em Proceedings of the 2018 Conference on Empirical Methods in
  Natural Language Processing}, pages 5039--5049. Association for Computational
  Linguistics.

\bibitem[Leng et~al., 2019]{leng2019unsupervised}
Leng, Y., Tan, X., Qin, T., Li, X.-Y., and Liu, T.-Y. (2019).
\newblock {Unsupervised Pivot Translation for Distant Languages}.
\newblock In {\em Proceedings of the 57th Annual Meeting of the Association for
  Computational Linguistics}, pages 175--183.

\bibitem[Li et~al., 2020]{li-etal-2020-reference}
Li, Z., Zhao, H., Wang, R., Utiyama, M., and Sumita, E. (2020).
\newblock Reference language based unsupervised neural machine translation.
\newblock In {\em Findings of the Association for Computational Linguistics:
  EMNLP 2020}, pages 4151--4162, Online. Association for Computational
  Linguistics.

\bibitem[Littell et~al., 2017]{littell-etal-2017-uriel}
Littell, P., Mortensen, D.~R., Lin, K., Kairis, K., Turner, C., and Levin, L.
  (2017).
\newblock {URIEL} and lang2vec: Representing languages as typological,
  geographical, and phylogenetic vectors.
\newblock In {\em Proceedings of the 15th Conference of the {E}uropean Chapter
  of the Association for Computational Linguistics: Volume 2, Short Papers},
  pages 8--14, Valencia, Spain. Association for Computational Linguistics.

\bibitem[Liu et~al., 2020]{liu2020mbart}
Liu, Y., Gu, J., Goyal, N., Li, X., Edunov, S., Ghazvininejad, M., Lewis, M.,
  and Zettlemoyer, L. (2020).
\newblock {Multilingual Denoising Pre-training for Neural Machine Translation}.
\newblock {\em Transactions of the Association for Computational Linguistics},
  8:726--742.

\bibitem[Marchisio et~al., 2020]{marchisio2020does}
Marchisio, K., Duh, K., and Koehn, P. (2020).
\newblock When does unsupervised machine translation work?
\newblock In {\em Proceedings of the Fifth Conference on Machine Translation},
  pages 571--583, Online. Association for Computational Linguistics.

\bibitem[Martinus and Abbott, 2019]{martinus2019focus}
Martinus, L. and Abbott, J.~Z. (2019).
\newblock {A Focus on Neural Machine Translation for African Languages}.
\newblock {\em CoRR}, abs/1906.05685.

\bibitem[McKellar and Puttkammer, 2020]{mckellar2020dataset}
McKellar, C.~A. and Puttkammer, M.~J. (2020).
\newblock {Dataset for comparable evaluation of machine translation between 11
  South African languages}.
\newblock {\em Data in Brief}, 29:105146.

\bibitem[Mikolov et~al., 2013]{mikolov2013distributed}
Mikolov, T., Sutskever, I., Chen, K., Corrado, G., and Dean, J. (2013).
\newblock Distributed representations of words and phrases and their
  compositionality.
\newblock In {\em Proceedings of the 26th International Conference on Neural
  Information Processing Systems - Volume 2}, NIPS'13, page 3111–3119, Red
  Hook, NY, USA. Curran Associates Inc.

\bibitem[Niu et~al., 2018]{niu2018bidirectional}
Niu, X., Denkowski, M., and Carpuat, M. (2018).
\newblock Bi-directional neural machine translation with synthetic parallel
  data.
\newblock In {\em Proceedings of the 2nd Workshop on Neural Machine Translation
  and Generation}, pages 84--91, Melbourne, Australia. Association for
  Computational Linguistics.

\bibitem[Papineni et~al., 2002]{papineni2002BLEU}
Papineni, K., Roukos, S., Ward, T., and Zhu, W.-J. (2002).
\newblock {BLEU: A Method for Automatic Evaluation of Machine Translation}.
\newblock In {\em Proceedings of the 40th Annual Meeting on Association for
  Computational Linguistics}, pages 311--318, Stroudsburg, PA, USA. Association
  for Computational Linguistics.

\bibitem[Post, 2018]{post-2018-call}
Post, M. (2018).
\newblock A call for clarity in reporting {BLEU} scores.
\newblock In {\em Proceedings of the Third Conference on Machine Translation:
  Research Papers}, pages 186--191, Belgium, Brussels. Association for
  Computational Linguistics.

\bibitem[Ramachandran et~al., 2017]{ramachandran-etal-2017-unsupervised}
Ramachandran, P., Liu, P., and Le, Q. (2017).
\newblock Unsupervised pretraining for sequence to sequence learning.
\newblock In {\em Proceedings of the 2017 Conference on Empirical Methods in
  Natural Language Processing}, pages 383--391, Copenhagen, Denmark.
  Association for Computational Linguistics.

\bibitem[Ren et~al., 2019]{ren2019unsupervised}
Ren, S., Zhang, Z., Liu, S., Zhou, M., and Ma, S. (2019).
\newblock {Unsupervised Neural Machine Translation with {SMT} as Posterior
  Regularization}.
\newblock In {\em The Thirty-Third {AAAI} Conference on Artificial
  Intelligence, {AAAI} 2019, Honolulu, Hawaii, USA}, pages 241--248. {AAAI}
  Press.

\bibitem[Ruiter et~al., 2019]{ruiter-etal-2019-self}
Ruiter, D., Espa{\~n}a-Bonet, C., and van Genabith, J. (2019).
\newblock {Self-Supervised Neural Machine Translation}.
\newblock In {\em Proceedings of the 57th Annual Meeting of the Association for
  Computational Linguistics}, pages 1828--1834, Florence, Italy. Association
  for Computational Linguistics.

\bibitem[Ruiter et~al., 2020]{ruiter2020selfinduced}
Ruiter, D., van Genabith, J., and Espa{\~{n}}a{-}Bonet, C. (2020).
\newblock {Self-Induced Curriculum Learning in Self-Supervised Neural Machine
  Translation}.
\newblock In {\em Proceedings of the 2020 Conference on Empirical Methods in
  Natural Language Processing (EMNLP)}, pages 2560--2571, Online. Association
  for Computational Linguistics.

\bibitem[Schwenk et~al., 2021]{schwenk2019wikimatrix}
Schwenk, H., Chaudhary, V., Sun, S., Gong, H., and Guzm{\'{a}}n, F. (2021).
\newblock {WikiMatrix: Mining 135M Parallel Sentences in 1620 Language Pairs
  from Wikipedia}.
\newblock In Merlo, P., Tiedemann, J., and Tsarfaty, R., editors, {\em
  Proceedings of the 16th Conference of the European Chapter of the Association
  for Computational Linguistics: Main Volume, {EACL} 2021, Online, April 19 -
  23, 2021}, pages 1351--1361. Association for Computational Linguistics.

\bibitem[Sen et~al., 2019]{sen-etal-2019-multilingual}
Sen, S., Gupta, K.~K., Ekbal, A., and Bhattacharyya, P. (2019).
\newblock Multilingual unsupervised {NMT} using shared encoder and
  language-specific decoders.
\newblock In {\em Proceedings of the 57th Annual Meeting of the Association for
  Computational Linguistics}, pages 3083--3089, Florence, Italy. Association
  for Computational Linguistics.

\bibitem[Sennrich et~al., 2016a]{sennrich-etal-2016-improving}
Sennrich, R., Haddow, B., and Birch, A. (2016a).
\newblock Improving neural machine translation models with monolingual data.
\newblock In {\em Proceedings of the 54th Annual Meeting of the Association for
  Computational Linguistics (Volume 1: Long Papers)}, pages 86--96, Berlin,
  Germany. Association for Computational Linguistics.

\bibitem[Sennrich et~al., 2016b]{sennrich-etal-2016-neural}
Sennrich, R., Haddow, B., and Birch, A. (2016b).
\newblock Neural machine translation of rare words with subword units.
\newblock In {\em Proceedings of the 54th Annual Meeting of the Association for
  Computational Linguistics (Volume 1: Long Papers)}, pages 1715--1725, Berlin,
  Germany. Association for Computational Linguistics.

\bibitem[ShweSin et~al., 2018]{yi2018myanmar}
ShweSin, Y.~M., Soe, K.~M., and Htwe, K.~Y. (2018).
\newblock {Large Scale Myanmar to English Neural Machine Translation System}.
\newblock In {\em 2018 IEEE 7th Global Conference on Consumer Electronics
  (GCCE)}, pages 464--465.

\bibitem[Tang et~al., 2020]{tang2020multilingual}
Tang, Y., Tran, C., Li, X., Chen, P.-J., Goyal, N., Chaudhary, V., Gu, J., and
  Fan, A. (2020).
\newblock Multilingual translation with extensible multilingual pretraining and
  finetuning.
\newblock {\em CoRR}, abs/2008.00401.

\bibitem[Yang et~al., 2018]{yangEtAl:2018}
Yang, Z., Chen, W., Wang, F., and Xu, B. (2018).
\newblock Unsupervised neural machine translation with weight sharing.
\newblock In {\em Proceedings of the 56th Annual Meeting of the Association for
  Computational Linguistics (Volume 1: Long Papers)}, pages 46--55. Association
  for Computational Linguistics.

\bibitem[Zhang et~al., 2018]{zhang2018joint}
Zhang, Z., Liu, S., Li, M., Zhou, M., and Chen, E. (2018).
\newblock Joint training for neural machine translation models with monolingual
  data.
\newblock In McIlraith, S.~A. and Weinberger, K.~Q., editors, {\em Proceedings
  of the Thirty-Second {AAAI} Conference on Artificial Intelligence, (AAAI-18),
  New Orleans, Louisiana, USA}, pages 555--562. {AAAI} Press.

\bibitem[Zoph et~al., 2016]{zoph-etal-2016-transfer}
Zoph, B., Yuret, D., May, J., and Knight, K. (2016).
\newblock Transfer learning for low-resource neural machine translation.
\newblock In {\em Proceedings of the 2016 Conference on Empirical Methods in
  Natural Language Processing}, pages 1568--1575, Austin, Texas. Association
  for Computational Linguistics.

\end{thebibliography}

\end{document}